\documentclass[numbered]{trbunofficial}
\usepackage{graphicx}

\usepackage[hidelinks]{hyperref}

\usepackage{booktabs}
\usepackage{subcaption}
\usepackage{multirow}
\usepackage{algorithm}
\usepackage{algorithmic}
\usepackage{mathrsfs}
\usepackage{amsfonts}
\usepackage{color}

\AuthorHeaders{Chen et al.}
\title{MetaFollower: Adaptable Personalized Autonomous Car Following}

\author{
    \textbf{Xianda Chen}\\
    Intelligent Transportation Thrust, Systems Hub\\
    The Hong Kong University of Science and Technology (Guangzhou), Guangzhou, 511400, China\\
    xchen595@connect.hkust-gz.edu.cn\\
    \hfill\break%
    \textbf{Kehua Chen}\\
    Division of Emerging Interdisciplinary Areas (EMIA), Interdisciplinary Programs Office\\
    The Hong Kong University of Science and Technology, Hong Kong, China\\
    kchenbm@connect.ust.hk\\
    \hfill\break%
    \textbf{Meixin Zhu, Ph.D., Corresponding Author}\\
     Assistant Professor\\
     Intelligent Transportation Thrust, Systems Hub\\
     The Hong Kong University of Science and Technology (Guangzhou), Guangzhou, 511400, China\\
     Guangdong Provincial Key Lab of Integrated Communication, Sensing and Computation for Ubiquitous Internet of Things, Guangzhou, 511400, China\\
    meixin@ust.hk\\
    \hfill\break%
     \textbf{Hao (Frank) Yang, Ph.D.}\\
     Research Assistant Professor\\
 Department of Civil and System Engineering, Whiting School of Engineering\\
 Johns Hopkins University, Baltimore, MD 21218\\
 haoya@jhu.edu\\
    \hfill\break%
     \textbf{Shaojie Shen, Ph.D. }\\
     Associate Professor \\
     Department of Electronic and Computer Engineering \\
     The Hong Kong University of Science and Technology, Hong Kong, China\\
     eeshaojie@ust.hk\\    
      \hfill\break%
    \textbf{Xuesong Wang, Ph.D.}\\
    Professor\\
    School of Transportation Engineering, Tongji University, Shanghai, 201804, China\\
    wangxs@tongji.edu.cn\\
    \hfill\break%
    \textbf{Yinhai Wang, Ph.D.}\\
    Professor\\
    Department of Civil and Environmental Engineering\\
    University of Washington, Seattle, WA 98195, USA\\
    yinhai@uw.edu  
}

\begin{document}
\maketitle

\section{Abstract}
Car-following (CF) modeling, a fundamental component in microscopic traffic simulation, has attracted increasing interest of researchers in the past decades. In this study, we propose an adaptable personalized car-following framework \textemdash\textendash \  
MetaFollower, by leveraging the power of meta-learning. Specifically, we first utilize Model-Agnostic Meta-Learning (MAML) to extract common driving knowledge from various CF events. Afterward, the pre-trained model can be fine-tuned on new drivers with only a few CF trajectories to achieve personalized CF adaptation. We additionally combine Long Short-Term Memory (LSTM) and Intelligent Driver Model (IDM) to reflect temporal heterogeneity with high interpretability. Unlike conventional adaptive cruise control (ACC) systems that rely on predefined settings and constant parameters without considering heterogeneous driving characteristics, MetaFollower can accurately capture and simulate the intricate dynamics of car-following behavior while considering the unique driving styles of individual drivers. We demonstrate the versatility and adaptability of MetaFollower by showcasing its ability to adapt to new drivers with limited training data quickly. To evaluate the performance of MetaFollower, we conduct rigorous experiments comparing it with both data-driven and physics-based models. The results reveal that our proposed framework outperforms baseline models in predicting car-following behavior with higher accuracy and safety. To the best of our knowledge, this is the first car-following model aiming to achieve fast adaptation by considering both driver and temporal heterogeneity based on meta-learning.



\hfill\break%
\noindent\textit{Keywords}: Car Following, Meta Learning, Driving Style Adaptation, Autonomous Driving, Adaptive Cruise Control. 
\newpage

\section{Introduction}
Car-following (CF) is a fundamental driving behavior in traffic flow, where each vehicle follows the one in front of it at a certain distance. Accurately modeling car-following behavior is crucial for microscopic traffic simulation and plays a key role in adaptive cruise control (ACC) systems. Over the years, researchers have shown significant interest in developing car-following models \cite{wang2016drivers, groelke2021predictive, wang2022effect, chen2023follownet, zhu2018modeling} to simulate and understand the dynamics of car-following.

Car-following models can be classified into three categories based on their modeling approaches \cite{zhang2022generative, mo2021physics, chen2024aggfollower}: physics-based models, data-driven models, and hybrid models. Physics-based models use heuristic rules and mathematical equations to simulate car-following behavior. Data-driven models, on the other hand, leverage large amounts of real-world driving data and machine learning techniques to extract patterns. Hybrid models combine both physics-based and data-driven approaches to take advantage of their respective strengths, aiming to improve the accuracy and interpretability of car-following models.

However, accurately capturing the heterogeneity in car-following behavior caused by different driving styles and temporal variations remains a challenge \cite{wang2018capturing, ossen2011heterogeneity, wang2018driving, xu2020driver, robbins2019does}. Previous research has identified several sources of heterogeneity in car-following behavior, such as age \cite{doroudgar2017driving}, gender \cite{kim2013driver}, vehicle type \cite{ravishankar2011vehicle}, road condition \cite{delitala2007mathematical}, as well as intra-driver heterogeneity \cite{chen2020investigating}. These pose challenges for accurately modeling car-following and lead to limitations on ACC systems. Most existing ACC systems utilize either a fixed time headway or distance headway setting, which lacks adaptability to individual drivers or varying driving conditions. Furthermore, developing personalized car-following models that can quickly learn from limited data and adapt to new drivers is a tricky problem and has hardly been investigated before.

\begin{figure*}[htbp]
\centering
\includegraphics[width=1\linewidth]{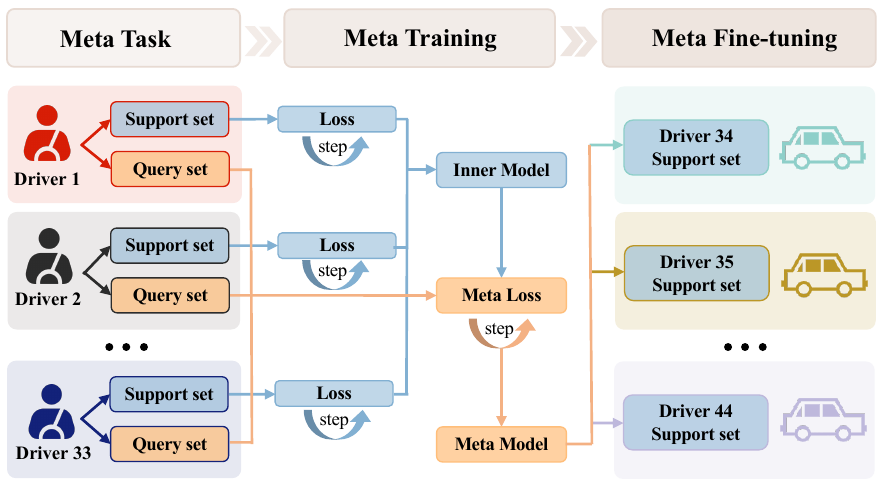}
\caption{A Roadmap of MetaFollower}
\label{fig:overview}
\end{figure*}

To address the aforementioned limitations, we propose MetaFollower, an adaptable personalized car-following framework that combines data-driven and physics-based models under the Model-Agnostic Meta-Learning (MAML) framework (Fig. \ref{fig:overview}). In detail, we leverage Long Short-Term Memory (LSTM) to generate parameters for the Intelligent Driver Model (IDM), incorporating temporal heterogeneity and interpretability. We then train the hybrid model with MAML to generate good initialization parameters, the pre-trained CF model can fast adapt to new drivers with only a few CF events under the MAML framework. To evaluate the performance of MetaFollower, we extracted 3050 driver-specific car-following events from the Shanghai Naturalistic Driving Study-world dataset, involving a total of 44 drivers. The results show that the proposed model outperformed both data-driven and physics-based models in terms of both accuracy and safety. 

The main contributions of this study are as follows:
\begin{itemize}
\item We introduce MetaFollower, the first car-following model based on meta-learning, that considers both driver and temporal heterogeneity. Our model enables fast adaptation to new drivers with limited training data.   
\item We visually segment the driving behavior of different drivers, showcasing the heterogeneity in driving behavior. 
\item We validate the effectiveness and superiority of our approach by conducting experiments using real-world driving data from naturalistic drivers and comparing it with baseline models.
\end{itemize}

\section{Related Work}

\subsection{Car-Following Model}
\subsubsection{Physics-Based Car-Following Models}

Physics-based car-following models utilize fundamental principles from traffic flow theory to describe the interaction between vehicles. These models incorporate parameters such as desired time headway, acceleration, and the effect of surrounding traffic conditions. Prominent examples in this category include Gazis-Herman-Rothery (GHR) model \cite{chandler1958traffic}, 
Helly's model \cite{helly1959simulation}, Wiedemann model \cite{wiedemann1974simulation}, Gipps Model  \cite{gipps1981behavioural}, Optimal Velocity Model (OVM) \cite{bando1995dynamical} and Intelligent Driver Model (IDM) \cite{treiber2000congested}. To gain a comprehensive understanding of conventional car-following models, \citet{chen2023follownet, brackstone1999car, saifuzzaman2014incorporating} provide valuable insights into the development and incorporation of car-following models in transportation research. These models are based on simplified assumptions and equations that may not fully capture the complexity and variability of real-world driving behavior. As a result, their accuracy may be lower compared to data-driven models.

\subsubsection{Data-Driven Car-Following Models}

Data-driven car-following models leverage large amounts of real-world driving data to establish models and extract patterns and rules with machine learning techniques. These models can learn from historical data and use statistical techniques to capture complex relationships between different variables, allowing for more accurate predictions and simulations compared to physics-based models. There are various data-driven car-following models available for predicting and modeling vehicle behavior in the car-following scenario. For example, Neural Networks (NN) \cite{hongfei2003develop, panwai2007neural, chong2013rule, khodayari2012modified, yang2018novel} are commonly used to capture complex relationships and make predictions based on observed data. LSTM \cite{zhou2017recurrent, cho2014learning, ma2020sequence, hochreiter1997long}, a type of recurrent neural network, is particularly suitable for handling time-series data and capturing temporal dependencies in car-following behavior. Transformer-based models, originally designed for natural language processing, can encode and decode vehicle trajectory data while considering global and local information for predicting vehicle behavior  \cite{zhu2022transfollower}. In addition, Reinforcement Learning (RL) \cite{zhu2018human, chai2015fuzzy, zhu2020safe, gao2018car,zhao2022personalized, song2023personalized} algorithms can train intelligent agents to learn tracking policies by providing states and rewards. However, data-driven approaches are black-box models and have low interpretability, which hinders their application.

\subsubsection{Hybrid Car-Following Models}

Hybrid car-following models, also known as physics-informed deep learning (PIDL) models, represent an innovative approach that combines the strengths of mathematical car-following models and data-driven methods. By incorporating physics-based constraints, these hybrid models enhance their interpretability. For example, \citet{yang2018novel} developed a combination car-following model that merges machine-learning-based and kinematics-based models by optimizing weight values. This model demonstrates superior performance in terms of safety and robustness compared to individual models. \citet{mo2021physics} proposed a PIDL approach for car-following models, combining physics-based models with deep-learning models. This integration enhances prediction accuracy and data efficiency by incorporating fundamental traffic flow theories. Building upon this approach, \citet{mo2022uncertainty} addressed uncertainty quantification in car-following behavior by integrating stochastic physics into the PIDL structure.

\subsection{Driving Heterogeneity in Car-Following Behavior}

Car-following behavior varies significantly among different drivers, reflecting the inherent heterogeneity in driving styles. Several studies have examined driving style heterogeneity in car-following behavior through field observations, driving simulator experiments, and data analysis. \citet{ossen2011heterogeneity} investigated the heterogeneity in car-following behavior using a large dataset of trajectory observations. It identifies significant differences in behavior among passenger car drivers and between passenger car drivers and truck drivers. This study also reveals that truck drivers exhibit a more consistent and robust car-following behavior compared to passenger car drivers. Research by \citet{ding2022driver} assumed that drivers share a set of driver states, and each driver has a unique driver profile that characterizes driving style. The method considers both intra-driver and inter-driver heterogeneity. \citet{wen2022characterizing} used the Waymo Open Dataset to investigate car-following behavior between human-driven vehicles (MVs) following automated vehicles (AVs) and MVs following MVs. Results show that MV-following-AV events have lower driving volatility, smaller time headways, and higher time to collision (TTC) values, and human drivers exhibit four distinct car-following styles. \citet{huang2022improving} presented an enhanced Fog-related Intelligent Driver Model (FIDM) that considers unobserved driver heterogeneity in fog conditions to accurately reproduce car-following behavior. The results demonstrated that as fog density decreased, unobserved driver heterogeneity increased. \citet{kim2013identifying} calibrated a car-following model with random coefficients, which can capture the heterogeneity across drivers who respond differently to stimuli. The expectation-maximization (EM) algorithm is employed to overcome challenges related to dimensionality and empirical identification. The calibration results confirm significant variations in random coefficients among drivers, with correlations between them.

\subsection{Meta-Learning}

Meta-learning, also known as "learning to learn," is a subfield of machine learning that focuses on developing algorithms and techniques that enable models to learn new tasks or adapt to new environments quickly and effectively. \citet{li2016learning} introduced a meta-learning approach for learning optimization algorithms. Through this framework, the authors successfully learn to optimize black-box functions using recurrent neural networks. This innovative application of meta-learning provides insights into the automatic design of optimization algorithms. Additionally, \citet{santoro2016meta} proposed a memory-augmented neural network architecture capable of learning new concepts from limited examples. The model employs external memory, resembling working memory in humans, to store information for generalization to unseen tasks. This work showcases the potential of meta-learning in enabling one-shot learning capabilities. \citet{finn2017model} proposed a Model-Agnostic Meta-Learning (MAML) algorithm that enables rapid adaptation of deep neural networks to new tasks. Their method demonstrates impressive performance across various domains, showcasing the potential of meta-learning techniques for efficient learning.  Furthermore, \citet{lee2018gradient} proposed a gradient-based meta-learning framework with learned layerwise metrics. The authors demonstrate improved performance on few-shot classification tasks by explicitly learning adaptive metrics within the network. Moreover, Meta-learning has applications in various domains, including computer vision \cite{chen2021meta}, natural language processing \cite{lee2022meta}, robotics \cite{alet2018modular}, and reinforcement learning \cite{schweighofer2003meta}. In regard to transportation, \citet{ye2021meta} proposed the use of Meta Reinforcement Learning (MRL) to improve the generalization capabilities of autonomous driving agents in new environments. Specifically, the method focuses on automated lane-changing maneuvers under different traffic congestion levels. Results show that the proposed MRL approach achieves higher success rates and lower collision rates compared to the benchmark model, particularly in heavy traffic conditions that were not encountered during training. \citet{jin2022selective} introduced CrossTReS, a framework for traffic prediction using meta-learning in transportation. It addresses the scarcity of data by utilizing cross-city transfer learning. Experimental results demonstrate that CrossTReS outperforms state-of-the-art baselines by up to 8\% in real-world scenarios.

\section{Preliminary Knowledge}

\subsection{Intelligent Driver Model (IDM)}
IDM is a widely recognized physics-based car-following model that considers both distance-based and time-based driving behavior. The model takes into account several influencing factors as input. These factors include the following vehicle's speed, the gap as well as the relative speed between the following vehicle (FV) and the lead vehicle (LV). It reproduces realistic car-following behavior, including smooth acceleration, maintaining a safe distance, and adapting to changes in speed and traffic conditions, the model expressions are:
\begin{equation}
a_n(t)=a_0\left[1-\left(\frac{v_n(t)}{\widetilde{v}_n}\right)^\lambda-\left(\frac{\widetilde{S}_n(t)}{s_n}\right)^2\right] 
\end{equation}
\begin{equation}
\widetilde{S}_n(t)=S_0+v_n(t) \widetilde{T}+\frac{v_n(t) \Delta v(t)}{2 \sqrt{a_0 b}} 
\end{equation}
\begin{equation}
\Delta v(t)=v_n(t)-v_{n-1}(t)
\end{equation}
where $a_n(t)$ and $v_n(t)$ represent the acceleration and the velocity of the FV at time $t$, respectively. $S_n(t)$, $\Delta v(t)$ are the spacing and relative speed between the FV and the LV. The desired maximum acceleration, comfortable deceleration, desired velocity, and desired time headway are represented by $a_0$, $b$, $\widetilde v$, and $\widetilde{T}$, respectively. $S_0$ is the minimum safe headway and $\lambda$ is a constant to be calibrated.

\subsection{Long Short Term Memory (LSTM) Networks}
\begin{figure*}[htbp]
\centering
\includegraphics[width=1\linewidth]{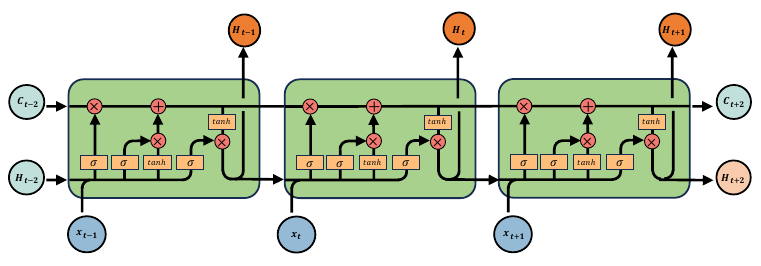}
\caption{LSTM Network Architecture \cite{hochreiter1997long}}
\label{fig:lstm}
\end{figure*}

LSTM \cite{hochreiter1997long} is a type of recurrent neural network (RNN) that addresses the vanishing gradient problem encountered in traditional RNNs. It is particularly useful for modeling car-following behavior due to its ability to capture temporal dependencies. Compared to a traditional RNN, LSTM has two main pathways: the cell state ($C$) and the hidden state ($H$). The cell state serves as the long-term memory, allowing important information to be preserved over time and preventing the vanishing gradient problem. The hidden state functions similarly to the hidden state in a regular RNN, capturing the short-term working memory. LSTM incorporates gates, such as the input gate, forget gate, and output gate, to control the flow of information. The input gate determines how much new information should be added to the cell state, while the forget gate controls what information should be discarded. The output gate controls which parts of the cell state should be outputted after analysis. The overall structure and flow of information in an LSTM model are illustrated in Fig. \ref{fig:lstm}. This architecture enables LSTM to effectively learn and predict car-following behavior by considering both short-term and long-term dependencies.

\subsection{Model-Agnostic Meta-Learning (MAML)}
MAML presents a novel and efficient meta-learning algorithm designed for the rapid adaptation of deep neural networks to new tasks using limited training data. The core idea of MAML is to learn an initial parameter $\theta$ that can be quickly adapted to new tasks using only a few gradient updates for each task. This adaptation process generates new parameters $\theta_i^{\prime}$ specific to each task. Subsequently, the global initial parameter $\theta$ is further updated from all tasks. Fig. \ref{fig:MAML} below illustrates this process, where the blue branching lines represent the update directions for different tasks, and the orange main axis line represents the overall direction of model parameters. This approach can be seen as a balance between different tasks, preventing parameters from overfitting to any individual task. The dashed line indicates the adaptation process for a new task, which involves fine-tuning model parameters. The adaptation process can be achieved with a few gradient updates, allowing the model to quickly adapt to the new task. The pseudocode for the MAML is shown in algorithm \ref{alg:MAML}.



\begin{figure}[htbp]
  \centering
  \includegraphics[width=1\textwidth]{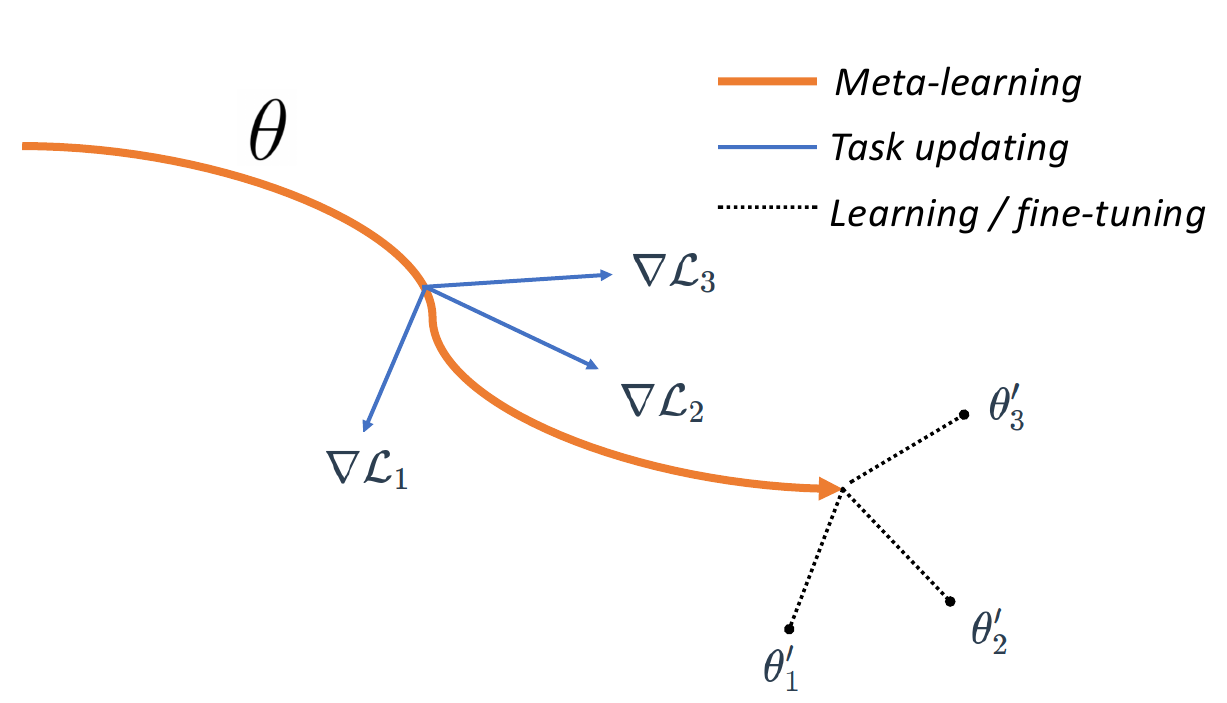}
  \caption{Illustration of the MAML Framework}
  \label{fig:MAML}
\end{figure}

\begin{algorithm}[htbp]
\caption{Model-Agnostic Meta-Learning (MAML)}
\label{alg:MAML}
\begin{algorithmic}[1] 
\REQUIRE{$p({T})$: distribution over tasks} 
\REQUIRE{$\alpha, \beta$: learning rates for inner and outer loop}
\REQUIRE{$K$: number of inner loop gradient steps}
\STATE Initialize $\boldsymbol{\theta}$: model parameters
\WHILE{not done} 
\STATE Sample batch of tasks ${T}_i \sim p({T})$
\FORALL{${T}_i$}
\STATE Sample $D_{\text{train}}^i$ and $D_{\text{test}}^i$  from ${T}_i$ 
\STATE Compute adapted parameters $\boldsymbol{\theta}_i^{\prime} \gets \boldsymbol{\theta} - \alpha \nabla_{\boldsymbol{\theta}}{L}_{{T}_i}(\boldsymbol{\theta}; D_{\text{train}}^i)$
\FOR{$k=1$ to $K$} 
\STATE Update adapted parameters $\boldsymbol{\theta}_i^{\prime} \gets \boldsymbol{\theta}_i^{\prime} - \alpha \nabla_{\boldsymbol{\theta}_i^{\prime}} {L}_{{T}_i}(\boldsymbol{\theta}_i^{\prime}; D_{\text{train}}^i)$ 
\ENDFOR 
\STATE Compute meta-objective ${L}_{\text{meta}}^i \gets {L}_{{T}_i}(\boldsymbol{\theta}_i^{\prime}; D_{\text{test}}^i)$ 
\ENDFOR
\STATE Update model parameters $\boldsymbol{\theta} \gets \boldsymbol{\theta} - \beta \nabla_{\boldsymbol{\theta}} \sum_{{T}_i \sim p({T})} {L}_{\text{meta}}^i$ \ENDWHILE 
\end{algorithmic} 
\end{algorithm}

\subsection{Problem Definition}

Given $H$ drivers with numerous car-following trajectories and $K$ new drivers with a few trajectories, we aim to train a model $ \mathcal{M} $ on $H$ drivers to extract driver styles and achieve fast adapting to the driving styles of the $K$ new drivers. Formally, we have a dataset $ {D}_H$ consisting of $H$ drivers and we treat each driver as one task under the MAML framework, where each driver $i$ is associated with a set of trajectories $ {T}_i = \{T_{i}^1, T_{i}^2, ..., T_{i}^N\}$, with $N$ being the number of trajectories for driver $i$. Each trajectory $T_{i}^n$ consists of a time sequence of states, represented as $(s_{m}^n, s_{m}^n, ..., s_{M}^n)$, where $M$ is the number of states in trajectory $T_{i}^n$. We also have a smaller dataset $ {D}_K$ containing trajectories from K new drivers, where each driver $j$ is associated with a set of trajectories $ {T}_{j} = \{T_{j}^1, T_{j}^2, ..., T_{j}^L\}$, with $L$ being the number of trajectories for driver $j$. Our objective is to train a model $ \mathcal{M} $ that can effectively extract driving styles from the trajectories in $ {D}_H$ and then adapt quickly to the driving styles in the trajectories of $ {D}_K$. To achieve this, the model $ \mathcal{M} $ needs to learn a mapping function $f$ that takes a trajectory $T_i^n$ as input and predicts the corresponding driving behavior. Mathematically, we can represent this as:
\begin{equation}
\hat{\bold y}_i^n = f(T; \theta)
\end{equation}
where $\hat{\bold y}_i^n$ is the predicted driving states of $n$ trajectory of driver $i$, $T$ is the trajectory, and $\theta$ represents the parameters of the model $ \mathcal{M} $. During training, the model $ \mathcal{M} $ learns the optimal values for the parameters $\theta$ by minimizing the discrepancy between the predicted driving behavior and the ground truth labels in the dataset $ {D}_H$. This can be formulated as an optimization problem:


\begin{equation}
\min_{\theta} \sum_{i=1}^{H} \sum_{n=1}^{N}  L(f(T_{i}^n; \theta), \bold y_{i}^n)
\end{equation}
where $L$ is the loss function that measures the discrepancy between the predicted behavior $f(T_{i}^n; \theta)$ and ground truth driving behavior $\bold y_{i}^n$. Once the model $ \mathcal{M} $ is trained on the dataset $ {D}_H$, it can be fine-tuned or adapted using the trajectories in $ {D}_K$. This adaptation process involves updating the parameters $\theta$ based on the trajectories and driving behavior of the new drivers. The objective is to minimize the discrepancy between the predicted $f(T_{j}^k; \theta)$ and true driving behavior $\bold y_{j}^k$ for the new drivers, similar to the training phase:

\begin{equation}
\min_{\theta}\sum_{j=1}^{K} \sum_{k=1}^{N} L(f(T_{j}^k; \theta), \bold y_{j}^k).
\end{equation}
By effectively training and adapting the model $ \mathcal{M} $ on both the dataset $ {D}_H$ and the trajectories from the new drivers in $ {D}_K$, we can achieve fast adaptation to the driving styles of the $K$ new drivers.

\section{Proposed Method}

In this work, we propose a novel model, MetaFollower, which combines the principle of meta-learning with a PIDL model. The PIDL model incorporates the IDM model into LSTM. The MetaFollower model is trained under the MAML framework.

\subsection{Driving Style Analysis}
\begin{figure*}[htbp]
    \centering
    \subfloat[Driver\ \textbf{2} Close Gap]{\includegraphics[width=0.32\linewidth]{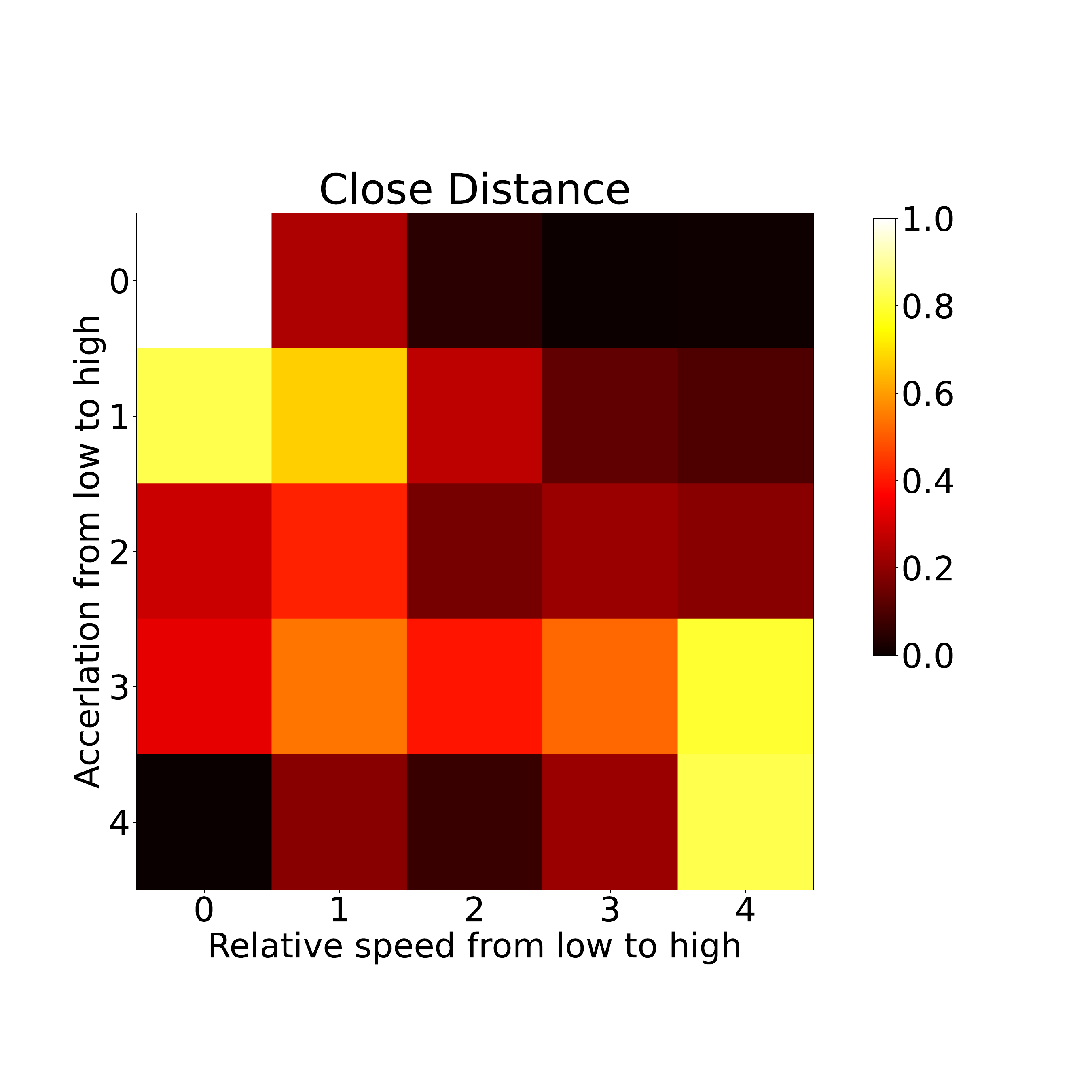}} 
    \subfloat[Driver\ \textbf{2} Normal Gap]{\includegraphics[width=0.32\linewidth]{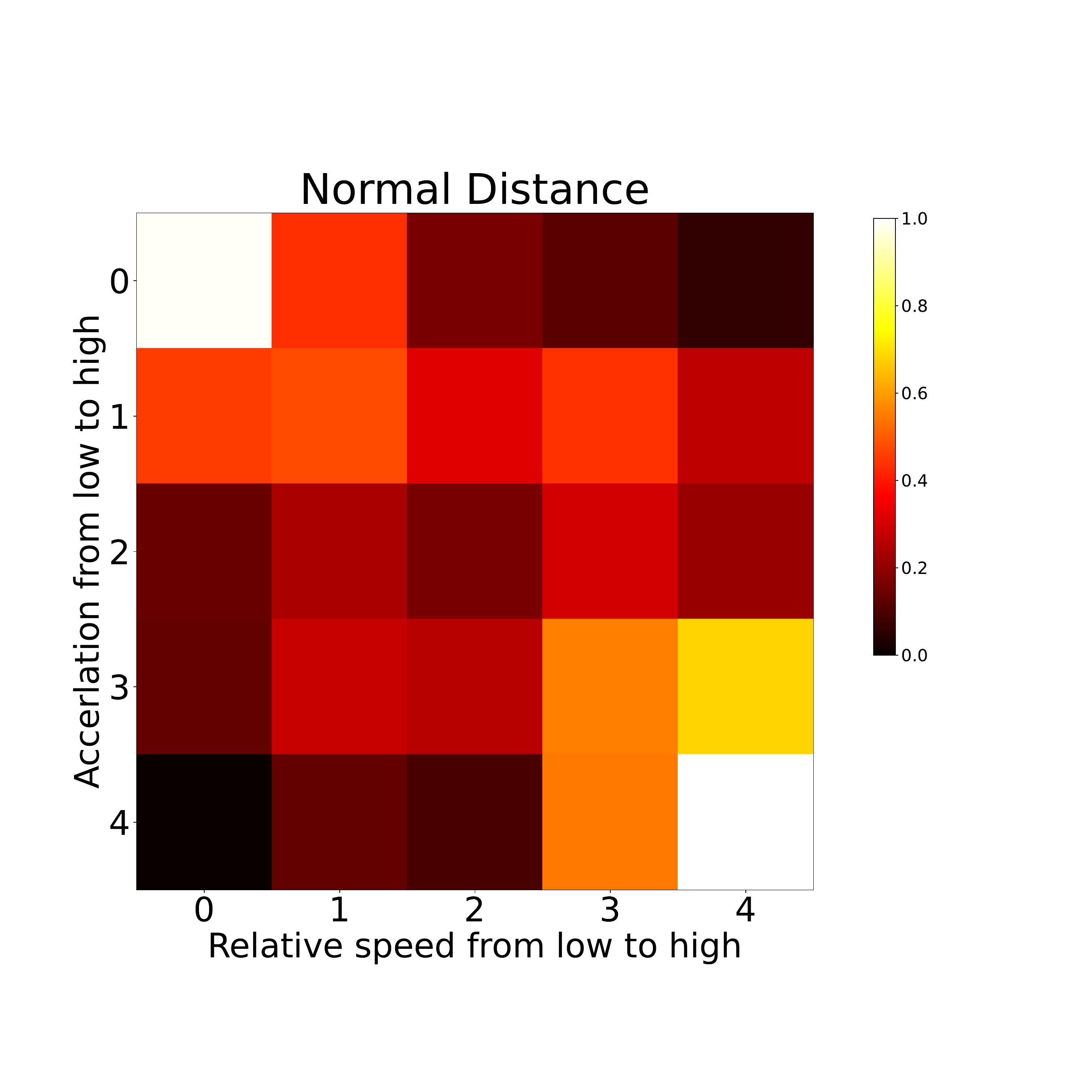}}
    \subfloat[Driver\ \textbf{2} Long Gap]{\includegraphics[width=0.32\linewidth]{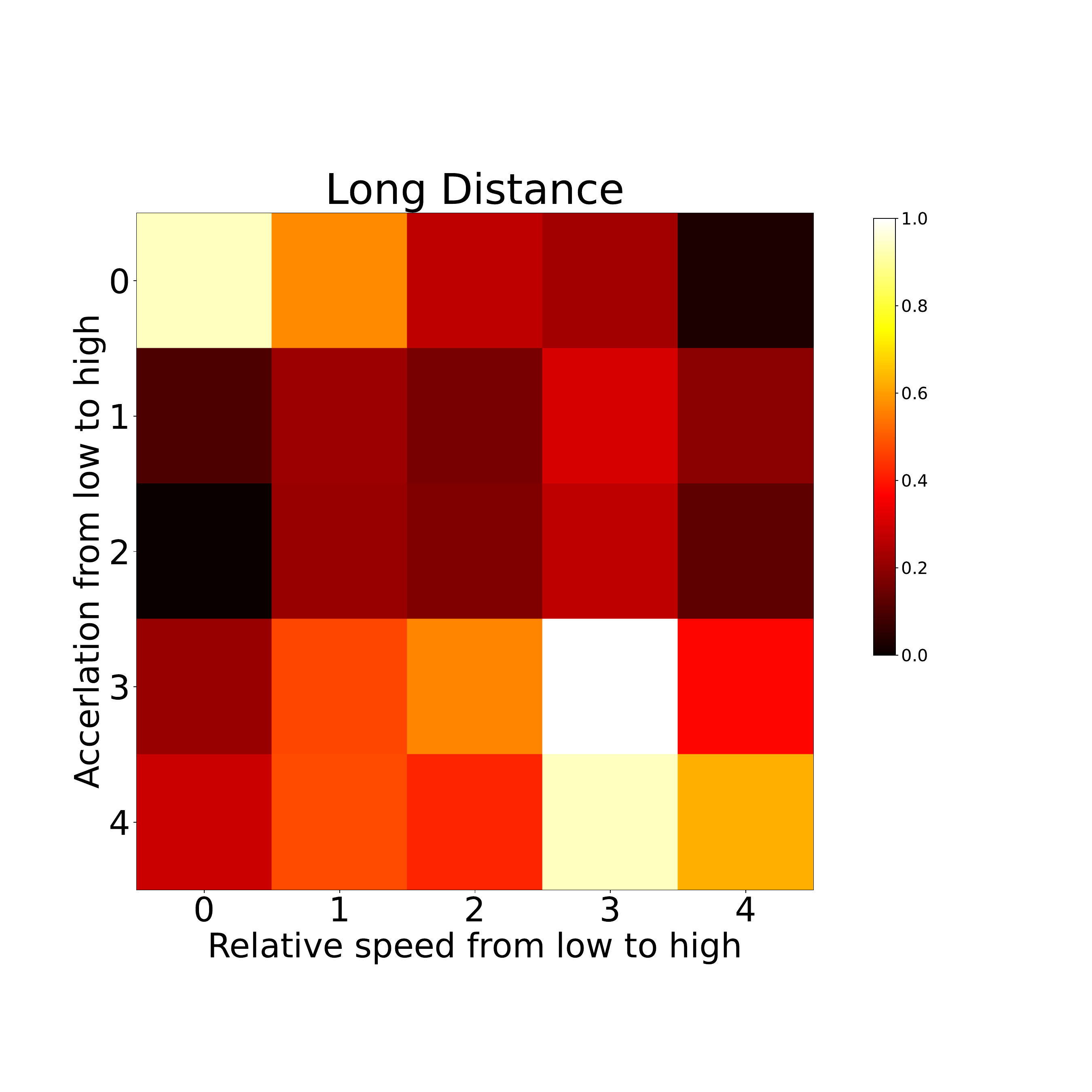}}\\
    \subfloat[Driver\ \textbf{8} Close Gap]{\includegraphics[width=0.32\linewidth]{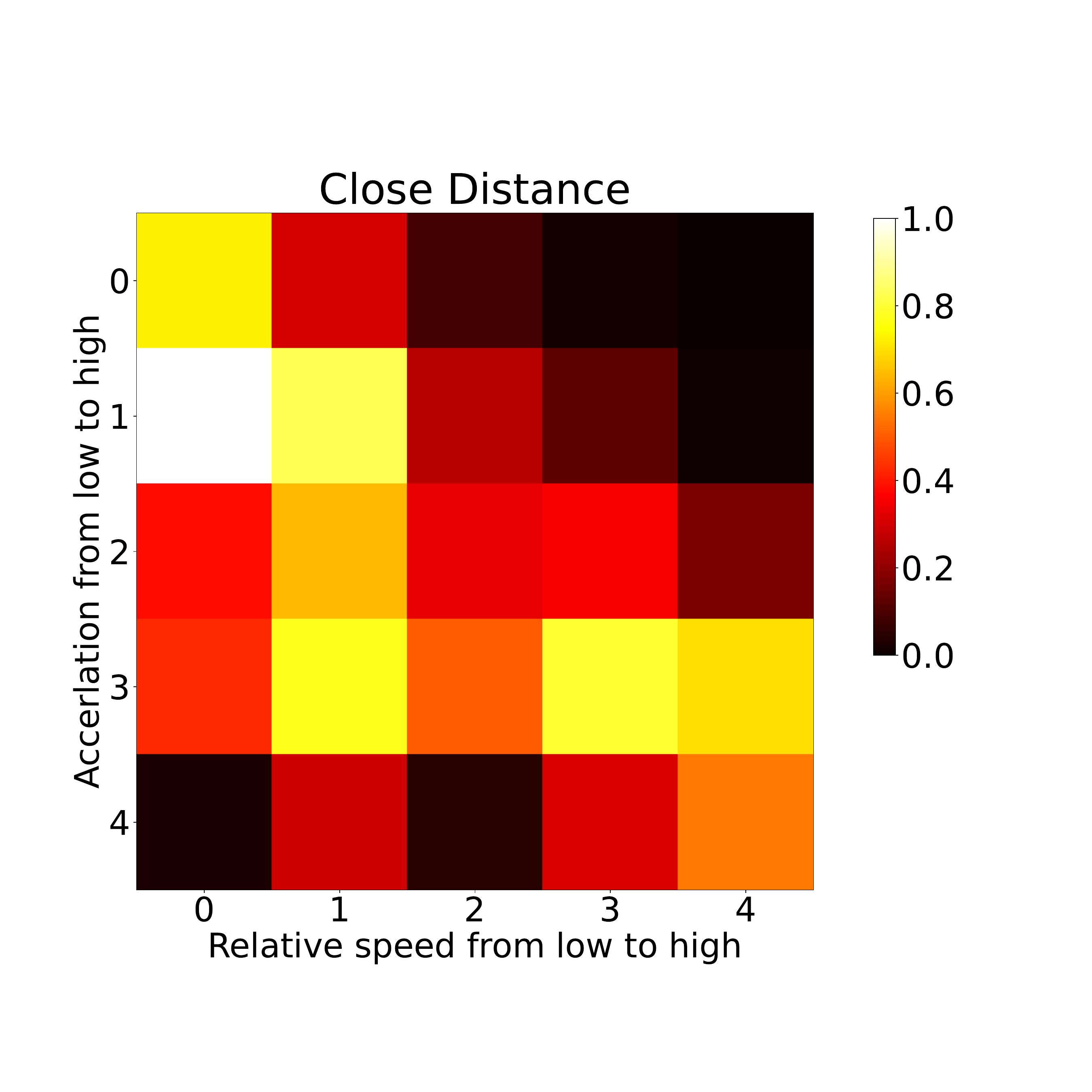}}
    \subfloat[Driver\ \textbf{8} Normal Gap]{\includegraphics[width=0.32\linewidth]{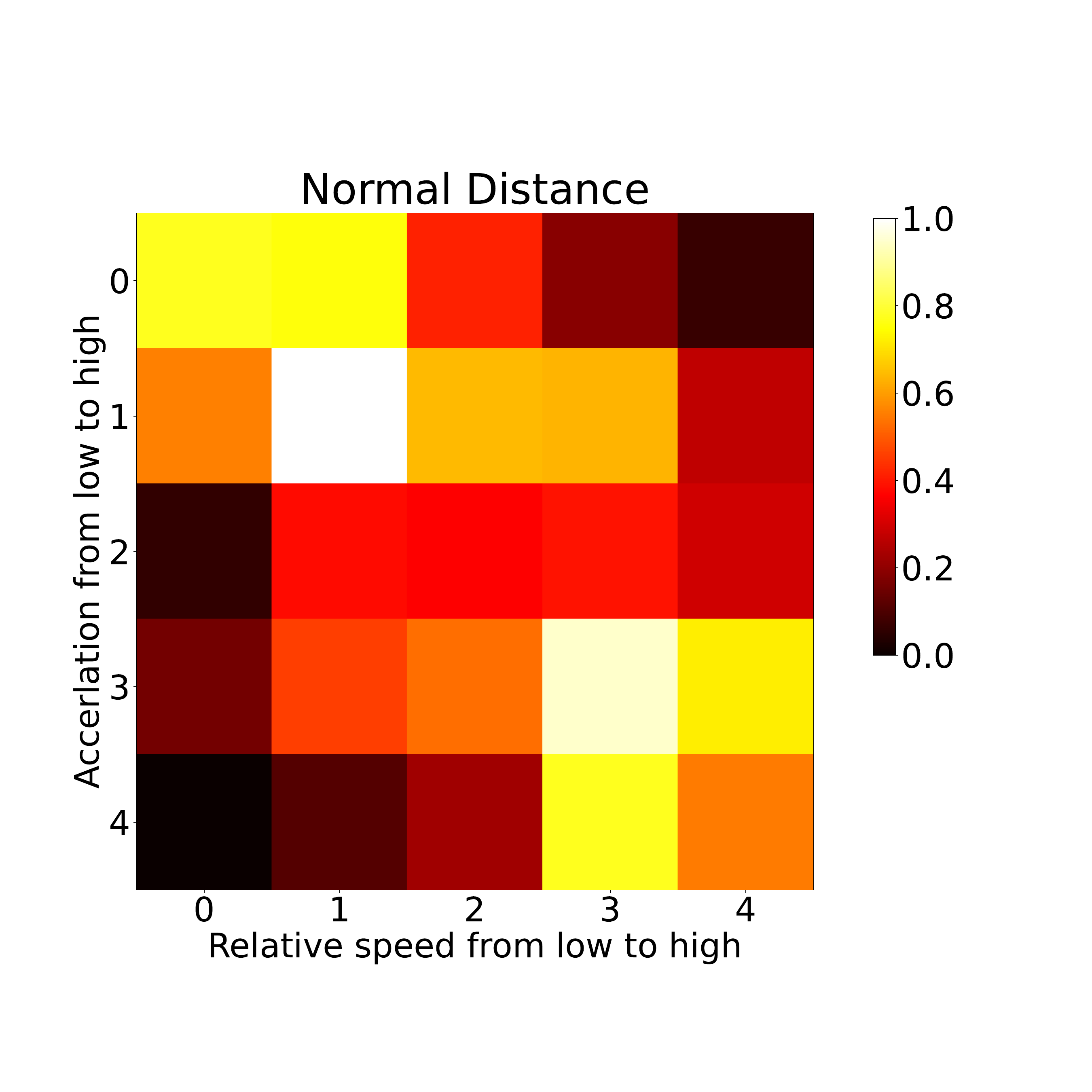}}
    \subfloat[Driver\ \textbf{8} Long Gap]{\includegraphics[width=0.32\linewidth]{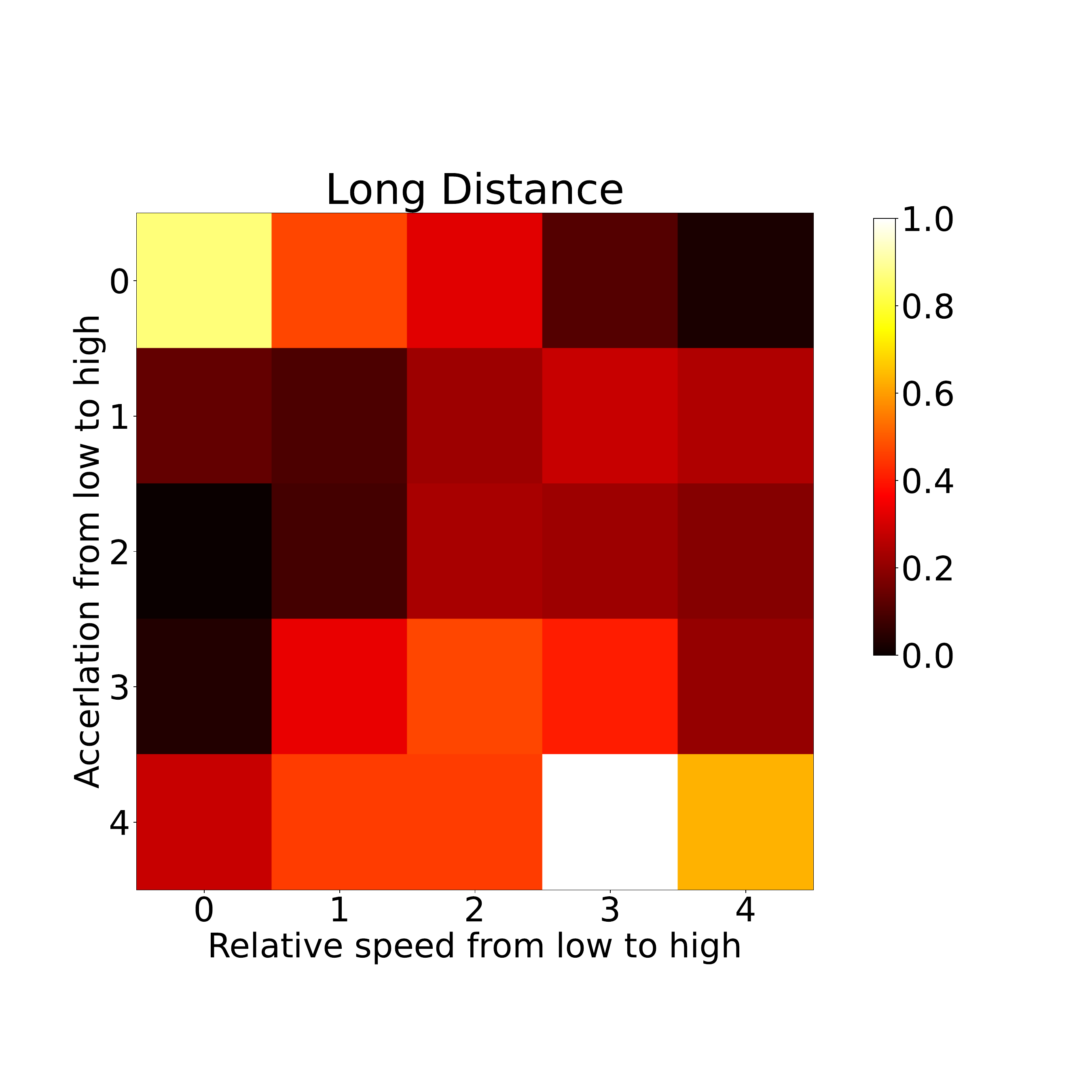}}\\
    \subfloat[Driver\ \textbf{10} Close Gap]{\includegraphics[width=0.32\linewidth]{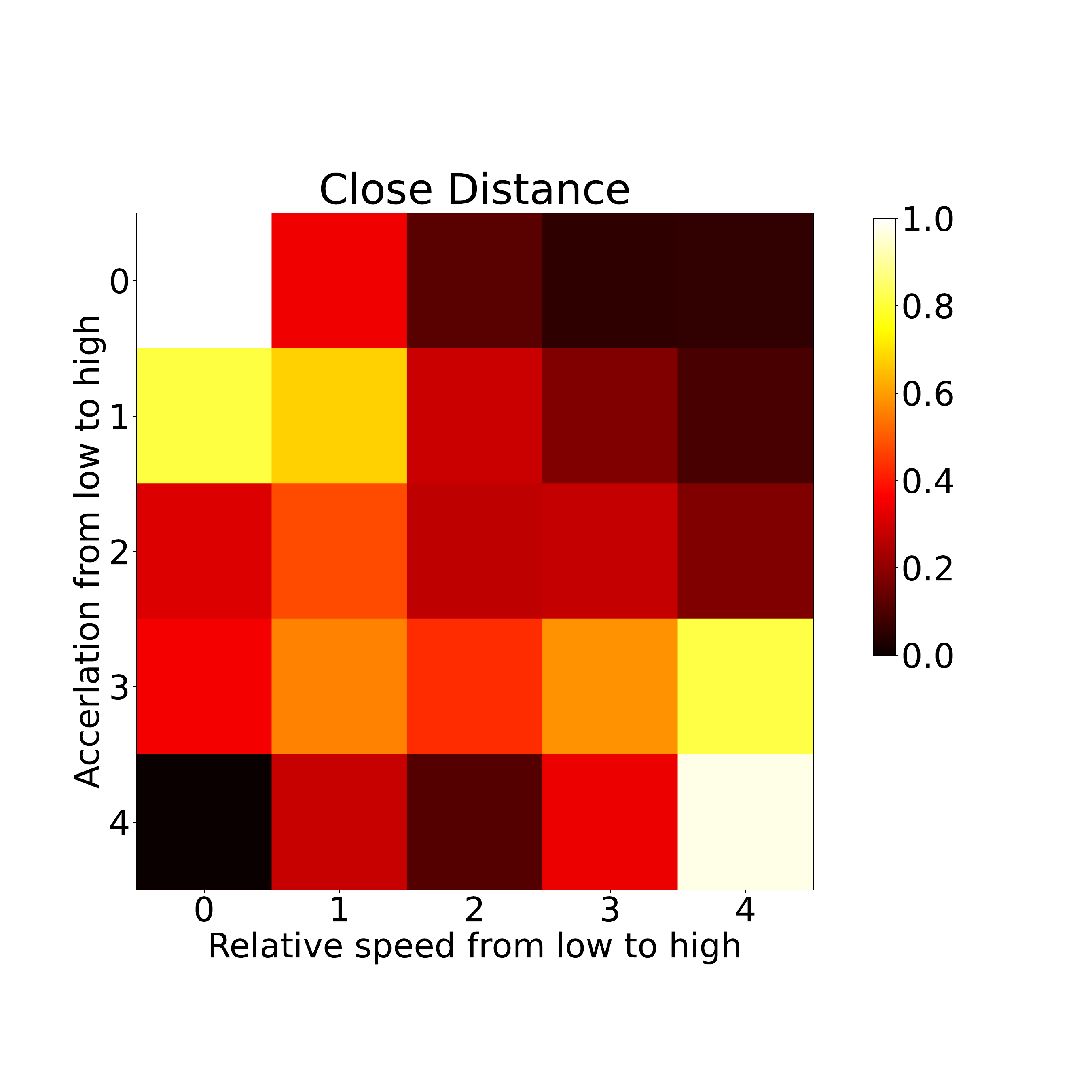}}
    \subfloat[Driver\ \textbf{10} Normal Gap]{\includegraphics[width=0.32\linewidth]{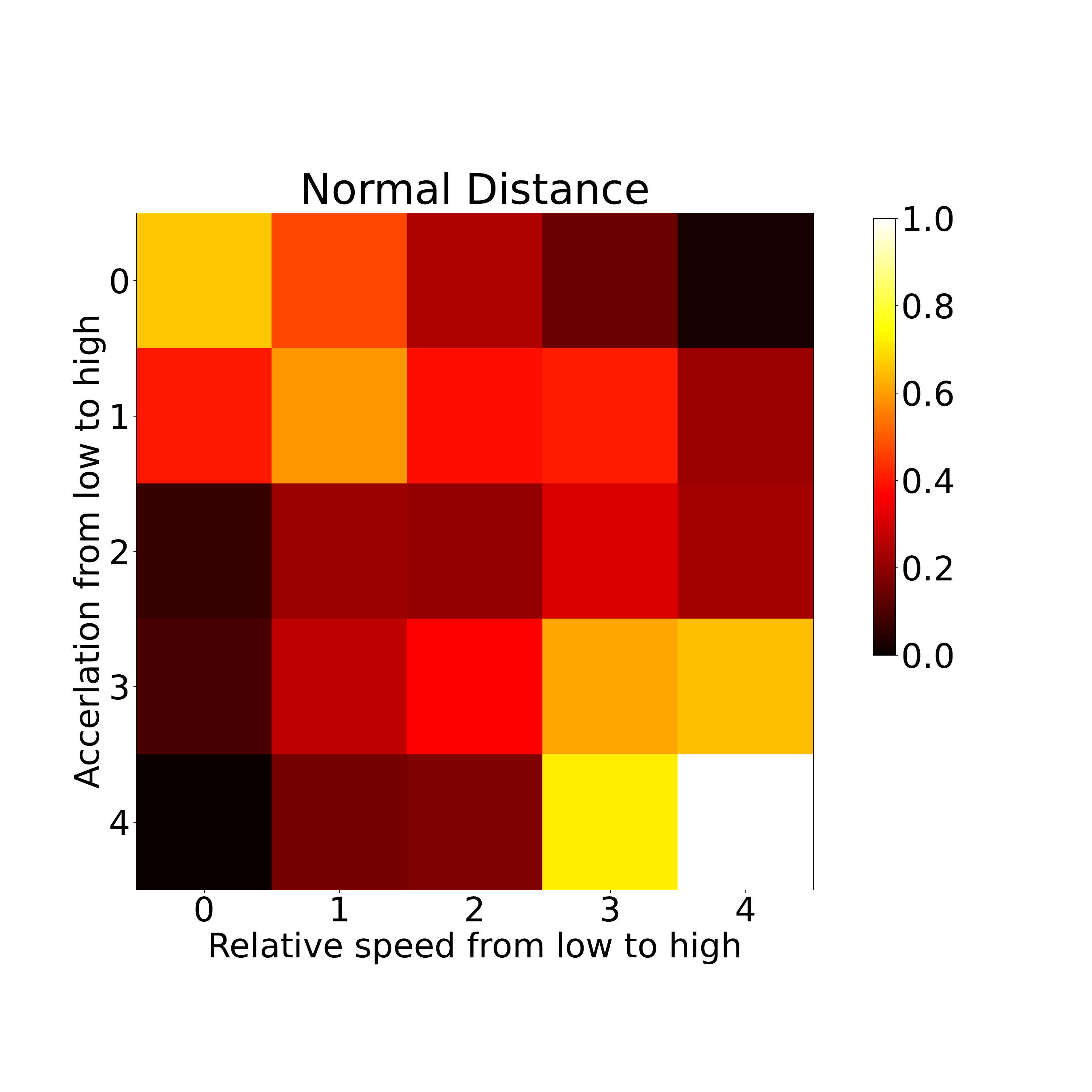}}
    \subfloat[Driver\ \textbf{10} Long Gap]{\includegraphics[width=0.32\linewidth]{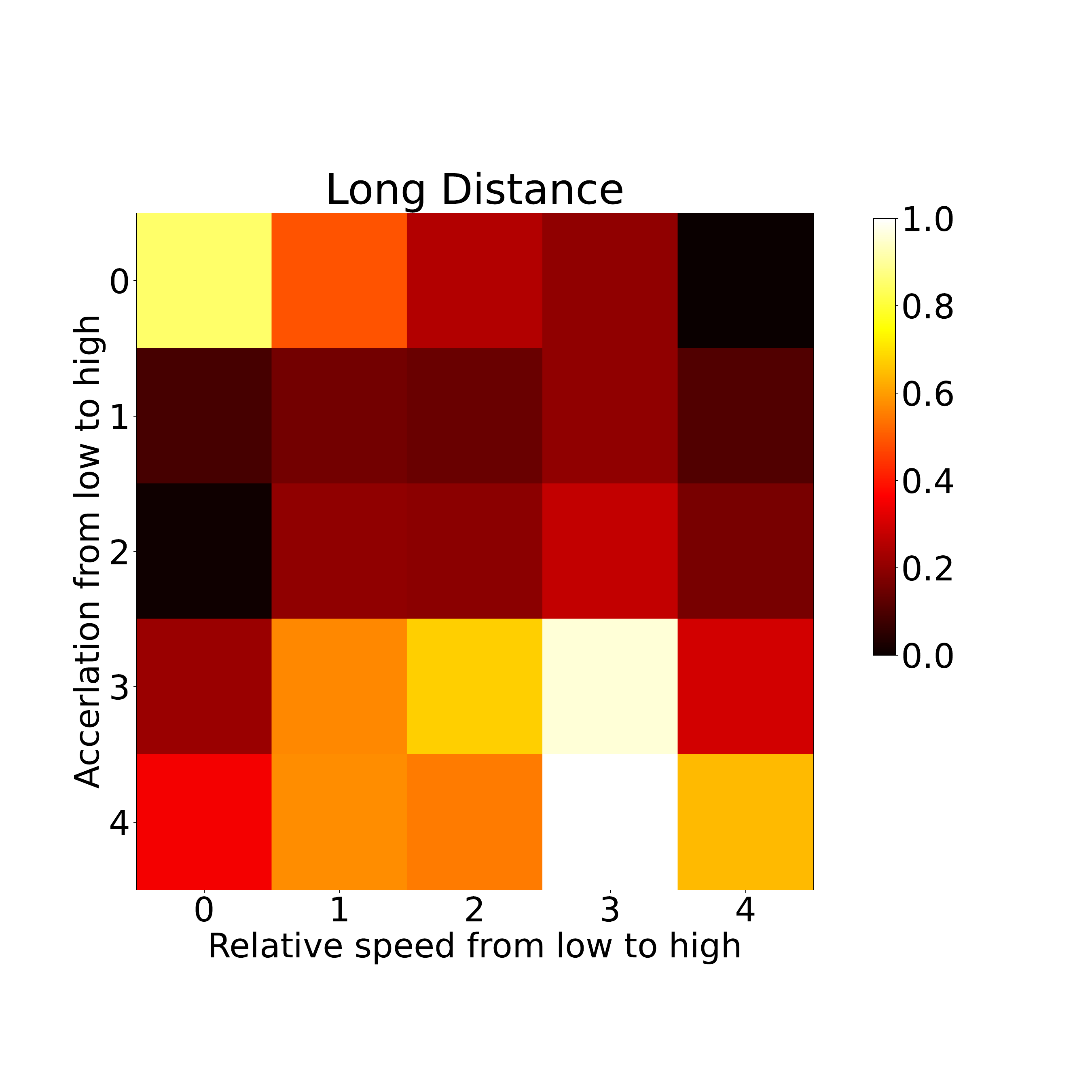}}    
    
    \caption{Distributions of Driving Patterns for Driver \textbf{2}, Driver \textbf{8} and Driver \textbf{10}}
        \label{fig:distribution}
\end{figure*}

\begin{figure*}[htbp]
    \centering
    \subfloat[Driver\ \textbf{2}]{\includegraphics[width=0.32\linewidth]{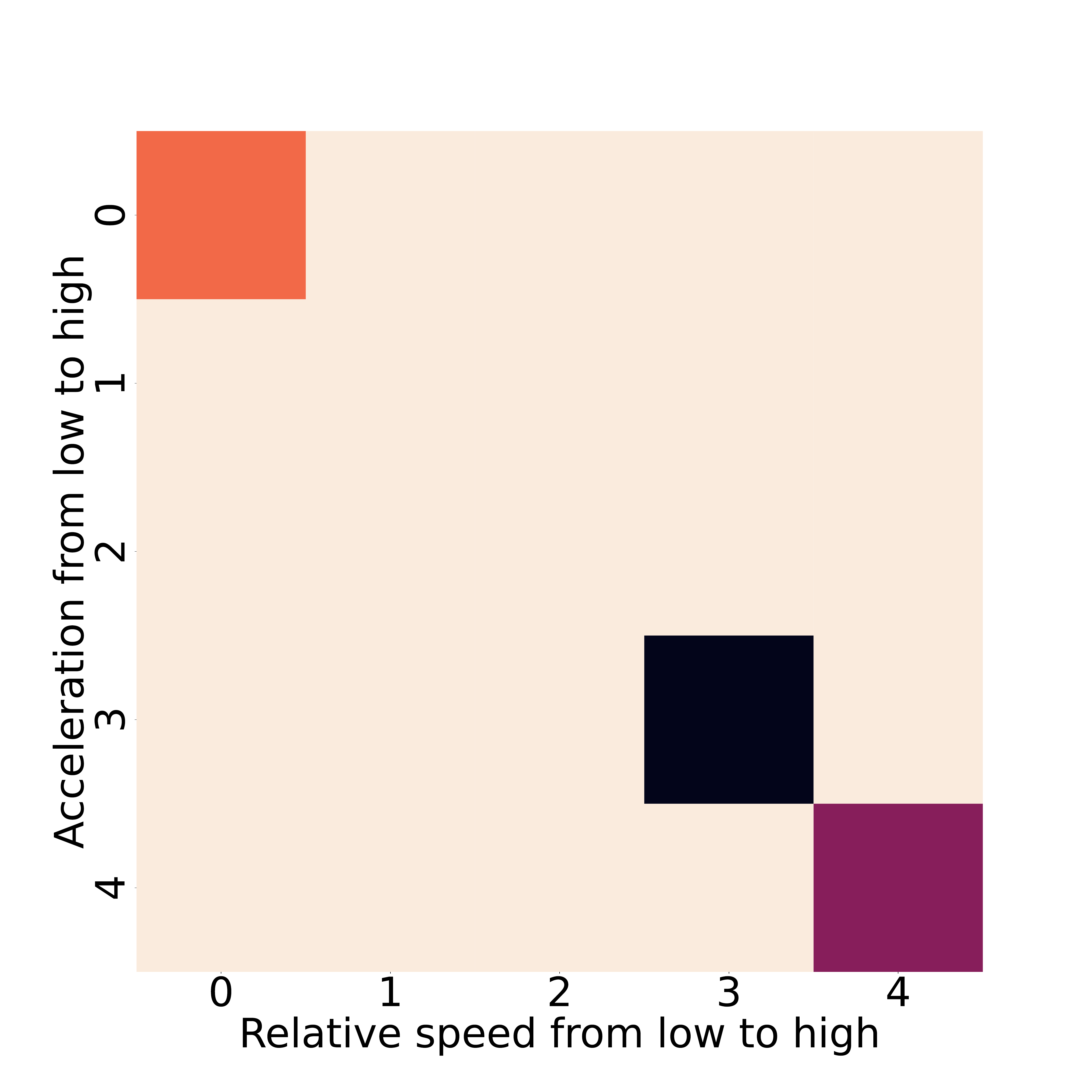}}
    \subfloat[Driver\ \textbf{8}]{\includegraphics[width=0.32\linewidth]{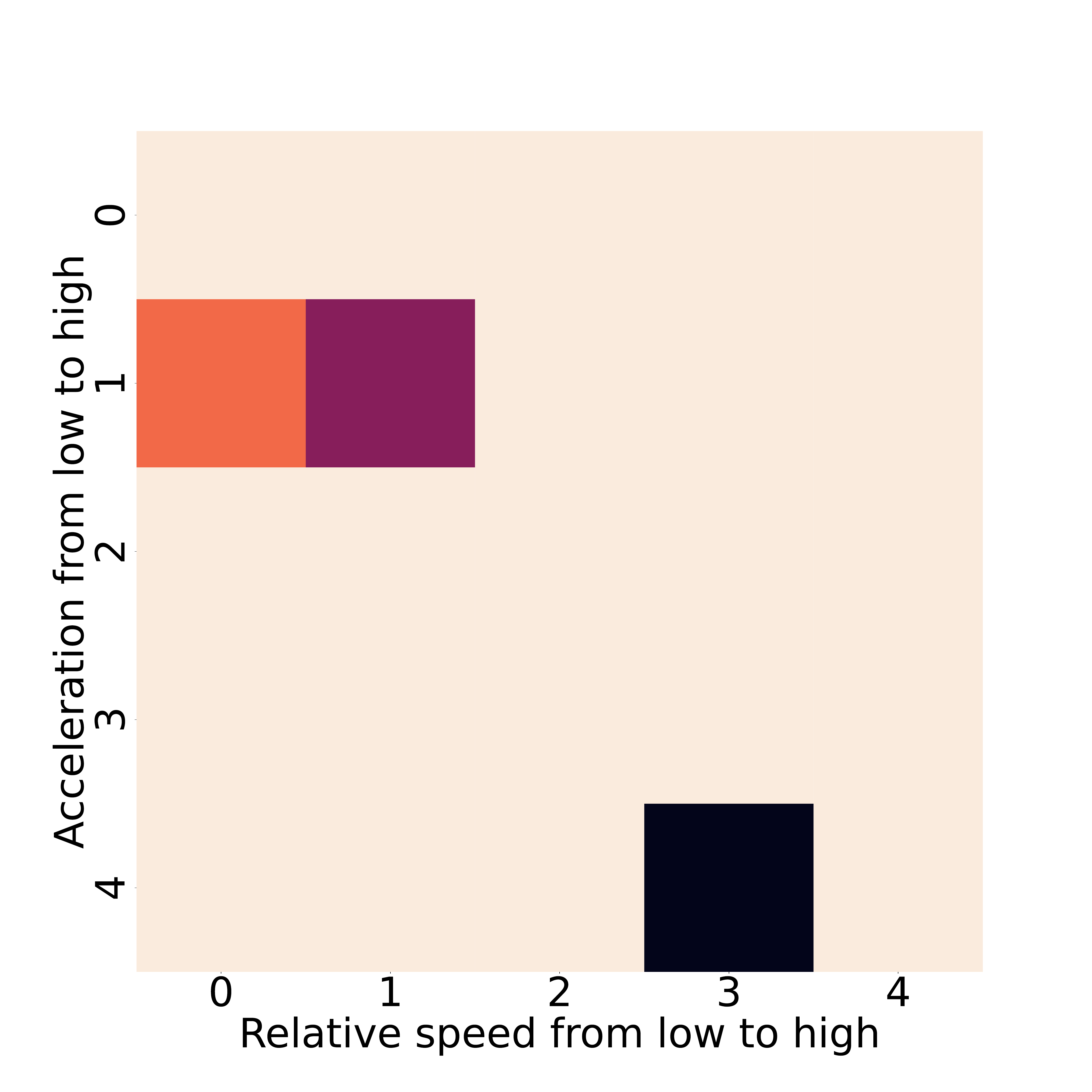}}
    \subfloat[Driver\ \textbf{10}]{\includegraphics[width=0.32\linewidth]{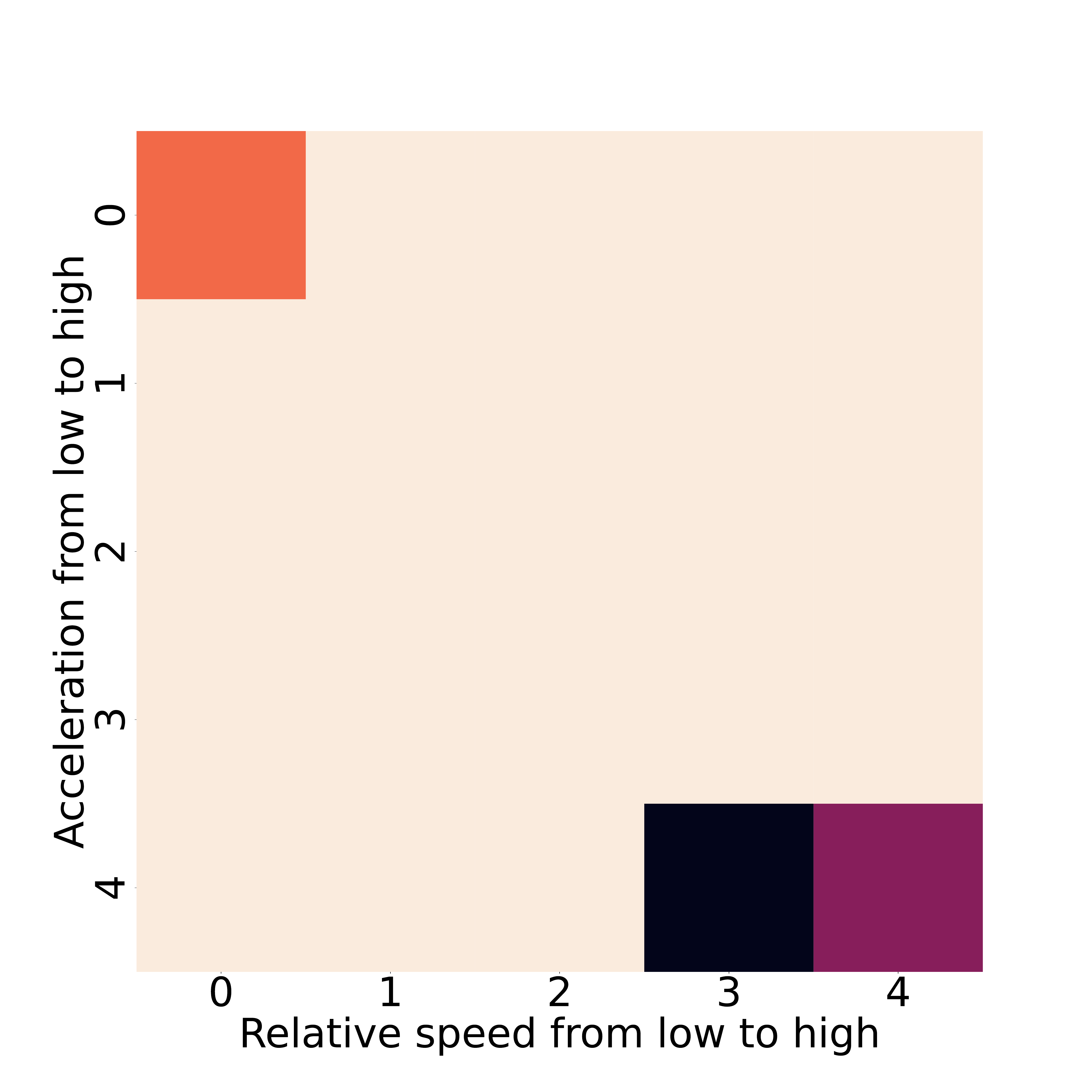}}\\ 
    \caption{Representative Driving Patterns for Driver \textbf{2}, Driver \textbf{8} and Driver \textbf{10}}
        \label{fig:distribution2}
\end{figure*}

To facilitate the semantic interpretation of the driving modes, we divided three variables into different levels based on the following driver's car-following data statistics \cite{wang2018driving}: spacing, relative speed, and acceleration. We fitted them using gamma distributions according to \cite{wang2018driving, bishop2006pattern} and determined the threshold for each variable from a statistical perspective, as shown in  Table \ref{tab:threshold}. Based on the predefined thresholds for each variable, we obtained a library of original car-following modes with a size of 75 (5 × 5 × 3  = 75). Fig. \ref{fig:distribution} illustrates the diverse driving behavior exhibited by three drivers under different car-following distances. The white color indicates a higher probability that the driver adheres to the corresponding mode, while the dark black color indicates a lower probability (close to zero). For instance, when following an LV at a long distance, driver 8 (Fig. \ref{fig:distribution}(f)) and driver 10 (Fig. \ref{fig:distribution}(i)) tend to approach the LV quickly, whereas driver 2 (Fig. \ref{fig:distribution}(c)) prefers a slower acceleration to approach the LV. To better visually differentiate the preferred driving modes of each driver, we selected the mode with the highest probability in each distance pattern, as shown in Fig. \ref{fig:distribution2}. The color orange represents the "close gap" mode, purple represents the "normal gap" mode, and black represents the "long gap" mode.

\begin{table*}[htbp]
\caption{Feature Distribution and Threshold Selection}
\centering
\label{tab:threshold}

\begin{tabular}{{@{}l|l|l@{}}}
\hline
\toprule
\textbf{Physical quantity} & \textbf{Threshold} & \textbf{Semantic state }          \\ \hline
\multirow{5}{*}{Acceleration {[}$m/s^2${]}} & (-$\infty$, -0.39)                        & Aggressive positive deceleration          \\ \cline{2-3} 
                                            & {[}-0.39, -0.08)                    & Gentle negative deceleration                      \\ \cline{2-3} 
                                            & {[}-0.08, 0.16)                     & Keeping acceleration                            \\ \cline{2-3} 
                                            & {[}0.16, 0.46)                      & Gentle positive acceleration                      \\ \cline{2-3} 
                                            & (0.46, +$\infty$)                         & Aggressive positive acceleration              \\ \hline
\multirow{5}{*}{Relative speed {[}$m/s${]}} & (-$\infty$, -0.82)                        & Aggressive negative relative speed        \\ \cline{2-3} 
                                            & {[}-0.82, -0.21)                    & Gentle negative relative speed                   \\ \cline{2-3} 
                                            & {[}-0.21, 0.28)                     & Keeping relative speed                           \\ \cline{2-3} 
                                            & {[}0.28, 0.89)                      & Gentle positive relative speed                  \\ \cline{2-3} 
                                            & (0.89, +$\infty$)                         & Aggressive positive relative speed             \\ \hline
\multirow{3}{*}{Spacing {[}$m${]}}          & (-$\infty$,10.11)                         & Close gap                                  \\ \cline{2-3} 
                                            & {[}10.11,24.70)                     & Normal gap                                 \\ \cline{2-3} 
                                            & (24.70,+$\infty$)                         & Long gap                          \\ 
\bottomrule
\end{tabular}
\end{table*}

\subsection{Improved PIDL Model with LSTM}

\begin{figure*}[htbp]
\centering
\includegraphics[width=1\linewidth]{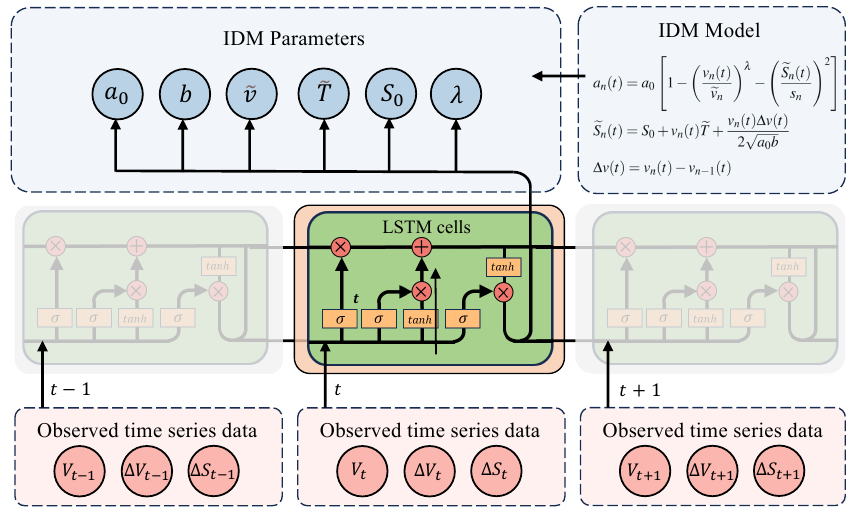}
\caption{LSTM-PIDL Model Architecture}
\label{fig:pinn}
\end{figure*}

In order to overcome the challenges associated with accurate and explainable predictions, our study presents an improved PIDL model that integrates the IDM into LSTM networks. According to \citet{zhu2018modeling}, IDM demonstrates superior performance compared to other traditional car-following models, making it the ideal choice as the physical component of our model. Additionally, \citet{mo2021physics} found that LSTM-PIDL outperforms NN-based PIDL models, which motivates us to select LSTM as the neural network component of our model. Unlike previous data-driven car-following models, which directly output the acceleration of the FV based on network parameters, the proposed PIDL model in this paper leverages LSTM networks to output the IDM model parameters. These parameters are then used with the IDM model to calculate the acceleration of the following vehicle. This approach essentially constructs a dynamic time-varying IDM model, providing greater flexibility and adaptability. This task can be formulated as an optimization problem:

\begin{equation}
\begin{aligned}\min_{\theta,\lambda}&&\sum_{i=1}^N\left(a^{(i)}-\widehat{a}^{(i)}\right)^2\\\mathrm{s.~t.}&&a^{(i)}=f_\theta\left({\mathbf{s}}^{(i)}|\theta\right),\quad i=1,\ldots,N,\\&&\psi\subseteq\Psi,\end{aligned}    
\end{equation}
where $\psi$ is the feasible domain of the parameters of the physics $\Psi$, representing the physical range of the IDM parameter, $\mathbf{s}^{(i)}$ is the $i$ th observed state, $f_\theta$ represents the PIDL model, and $\widehat{a}$ represents the predicted acceleration. The model architecture is shown as Fig. \ref{fig:pinn}. 

\subsection{MAML Updating}
In MAML, the support set and query set are two essential components used in the meta-learning process. The support set refers to a small labeled dataset that is provided to the model during the meta-training phase. It contains examples from various tasks or domains that the model needs to learn from. For each task, the support set includes both input data and corresponding target labels. The query set, on the other hand, is another dataset used for evaluating the model's performance after it has been trained on the support set. The query set usually consists of unseen examples from the same tasks or domains encountered in the support set. The MAML algorithm is employed to update the base learner, which is the improved PIDL model with IDM. This process involves two key stages: the inner-updating and the outer-updating.

\begin{itemize} \item \textbf{Inner-updating:} In the inner loop, model parameters are updated for each task using a small dataset (support set) from that task. The inner-updating for the PIDL model is performed using gradient descent and the task-specific loss function. The inner loop adaptation is performed as below:

\begin{equation}
    \boldsymbol{\theta}_i^{\prime} \gets \boldsymbol{\theta} - \alpha \nabla_{\boldsymbol{\theta}} {L}_{{T}_i}(\boldsymbol{\theta}; D_{\text{train}}^i),
\end{equation}
where $\boldsymbol{\theta}$ represents initial model parameters, $\alpha$ is the inner loop learning rate, and ${L}_{{T}_i}$ is the task-specific loss function for task ${T}_i$.

\item \textbf{Outer-updating:} In the outer loop, initial model parameters are updated based on the performance of the adapted model on the query set for each task. The meta-objective is computed as the average loss across all tasks. The outer loop adaptation is performed as below:

\begin{equation}
    \boldsymbol{\theta} \gets \boldsymbol{\theta} - \beta \nabla_{\boldsymbol{\theta}} \sum {L}_{\text{meta}}^i,
\end{equation}
where $\beta$ is the outer loop learning rate, and ${L}_{\text{meta}}^i$ is the meta loss for task ${T}_i$ computed from the adapted model parameters $\boldsymbol{\theta}_i^{\prime}$.
\end{itemize}

\subsection{Fine-tuning with MAML}

By leveraging the MAML algorithm, we obtain a pre-trained MetaFollower model that captures driving styles from the training set of drivers. This pre-trained model can then be fine-tuned based on a limited amount of data from new drivers to obtain the final car-following model, named MetaFollower. The fine-tuning process is outlined in the following steps:

\begin{enumerate} \item Initialize model parameters with the learned initialization from the MAML training: $\boldsymbol{\theta}^* \gets \boldsymbol{\theta}$. \item For each new driver $k \in {K}$, sample a small dataset (support set) for training. \item Perform the inner loop adaptation for each new driver using the support set and the learned initialization $\boldsymbol{\theta}^*$:

\begin{equation}
    \boldsymbol{\theta}_k^{\prime} \gets \boldsymbol{\theta}^* - \alpha \nabla_{\boldsymbol{\theta}^*} {L}_{{T}_k}(\boldsymbol{\theta}^*; D_{\text{train}}^k).
\end{equation}

\item Evaluate the adapted model's performance on the query set for each new driver and assess the model's ability to generalize to new driving styles with limited data.
\end{enumerate}
The fine-tuning process enables the MetaFollower model to adapt quickly to the driving styles of the $K$ new drivers, effectively capturing their car-following behavior with minimal additional training data.

\section{Data and Experiments}

\subsection{Shanghai Naturalistic Driving Study (SH-NDS)}
\begin{figure}[!h]
\centering
\includegraphics[width=0.8\linewidth]{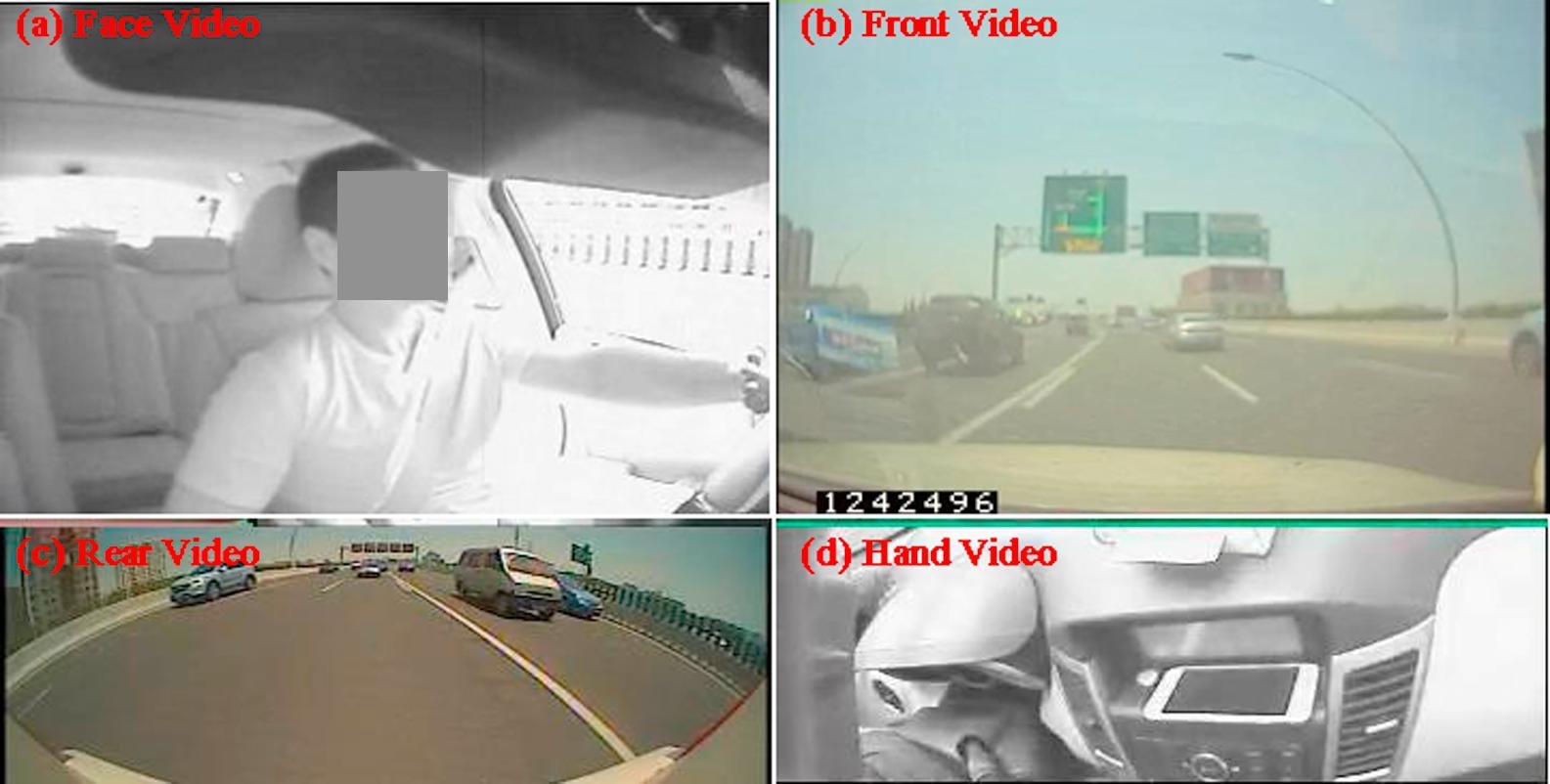}
\caption{SH-NDS Four Camera Views \cite{zhu2018human}}
\label{fig:shnds}
\end{figure}

This study utilized real-world car-following events from the Shanghai Naturalistic Driving Study (SH-NDS) to train and evaluate the proposed model. The SH-NDS, a collaborative effort by Tongji University, General Motors, and the Virginia Tech Transportation Institute, aimed to gain a deeper understanding of Chinese drivers' vehicle usage, operation, and safety awareness. The data collection spanned from December 2012 to December 2015, during which five passenger vehicles equipped with the second Strategic Highway Research Program (SHRP 2) Driving Automation System (DAS) \cite{dingus2015naturalistic} were employed. The DAS system encompassed a multitude of components, including a forward radar for measuring distance and relative speed to vehicles ahead, an accelerometer for tracking longitudinal and lateral acceleration, a GPS sensor for precise location information, an interface box for collecting vehicle CAN Bus data, and four synchronized video cameras that captured crucial aspects such as the driver's face, the view of the road ahead, the rear roadway, and the driver's hand movements, as illustrated in Fig.\ref{fig:shnds}.

\subsection{Car-Following Event Extraction}
Specific criteria were applied to extract car-following data from the SH-NDS dataset. These criteria were based on previous studies conducted by \citet{zhu2018human, wang2017capturing, zhao2017trafficnet}. The following criteria were used:
\begin{itemize}

\item The identification number of the LV remained constant, indicating that the FV was consistently following the same vehicle.

\item The lateral distance between the LV and the LV was less than 2.5 meters, ensuring that they were driving in the same lane.

\item The duration of the car-following event was longer than 15 seconds, providing sufficient data for analysis.

\item Each individual driver's dataset should have a minimum of 20 car-following events. This requirement ensured an adequate number of data samples for each driver, allowing for reliable statistical analysis and better capturing of individual driving behavior characteristics during the modeling. 
\end{itemize}
Based on these criteria, 44 Shanghai drivers were selected for car-following model research. In our study, we regard the car following data of each individual driver as separate MAML tasks. We divide the data of 44 drivers into a training set and a test set. Specifically, the data of 33 drivers were used as the training data, while the remaining 11 drivers' data were used as the test data. In the training dataset of the 33 drivers, each driver's data was further divided into a support set (33.3\% of the data) and a query set (66.7\% of the data). For the 11 test drivers, each driver's data was also divided into a support set (33.3\% of the data) and a query set (66.7\% of the data). In other words, the ratio of the support set to the query set is 1:3.
\subsection{Baselines}
Six types of baseline models are used for comparison to demonstrate the performance of our proposed model. All models use consistent input parameters, including spacing, FV's speed, and relative speed. The primary objective of these models is to predict the acceleration of the FV based on these inputs.

\begin{itemize}
\item \textbf{IDM}: The IDM model has been shown to outperform other traditional models according to \cite{zhu2018modeling}. We trained an IDM model using the genetic algorithm (GA) to minimize the Mean Squared Error (MSE) of spacing. By employing GA, we determined the optimal set of IDM parameters.

\item \textbf{GHR}: The GHR model is based on the assumption that the acceleration of the LV depends on the comparison between its own speed and that of the preceding vehicle. Similar to the IDM model, we employed GA to identify the optimal parameters for the GHR model \cite{zhu2018modeling}.

\item \textbf{LSTM without PIDL without meta without pre-train}:
This baseline model is trained directly on the training data of the test drivers without any pre-training or meta-learning. It uses the LSTM architecture to model the sequence data and predicts the acceleration of the FV based on the input parameters. 

\item \textbf{LSTM without PIDL without meta}:
In this model, pre-training is performed on the training drivers' data using the LSTM architecture. The model is then fine-tuned on the support set of the test drivers to adapt it to the specific characteristics of the test drivers' data. The model does not utilize meta-learning techniques and relies solely on the LSTM architecture for prediction.

\item \textbf{LSTM without PIDL with meta}:
This model combines the LSTM architecture with meta-learning techniques. In this case, the model is trained on the training drivers' data using meta-learning, which helps it adapt and generalize better to the new drivers' data in the support set.

\item \textbf{LSTM with PIDL without meta}:
This model combines the designed PIDL with the LSTM architecture. This  allows the model to leverage both the temporal patterns in the data and the prior knowledge of the underlying physics. Unlike the previous model, it does not utilize meta-learning techniques.
\end{itemize}

\subsection{Test Details}
According to the general meta-learning data splitting method, as mentioned before, we divide the drivers into training drivers (numbered 1 to 33) and testing drivers (numbered 34 to 44). Each driver has its own support set and query set, and all models are evaluated on the query set of the testing drivers.

\section{Results}

\subsection{Evaluation Metrics}
To evaluate the performance of a car-following model, we choose two metrics \cite{chen2023follownet} as the standard evaluation criteria: MSE of spacing and collision rate. The MSE of spacing measures the accuracy of the model in predicting the spacing between vehicles. A lower MSE score indicates a better fit of the model to the data and improved precision in predicting the spacing. For one car-following event, the MSE of spacing can be expressed as:

\begin{equation}
    \operatorname{MSE}={\frac{1}{N} \sum_{i=1}^{N}\left({{S_{n - 1,n}}(t) - S_{n - 1,n}^{obs}(t)}\right)^{2}}
\end{equation}
where $N$ is the total number of observations, and $i$ is an observation index. $S$ and $S^{obs}$ are the predicted and observed spacing between the FV and LV, respectively. 
Similarly, we define collision rate as the number of car-following events where the spacing between vehicles is less than zero divided by the total number of car-following events in the test dataset. The formula for collision rate is as follows:

\begin{equation}
Collision\ Rate = \frac{\text{Number of events with spacing} < 0}{\text{Total number of car-following events}}
\end{equation}

\subsection{MSE and Collision Rate in Testing Dataset}
According to the analysis of Table \ref{tab:results}, we can observe the performance of different models in terms of MSE of spacing and collision rate. Below is the analysis for each model:
\begin{itemize}
\item GHR Model: It shows poor performance in spacing prediction but has a good collision avoidance rate.

\item IDM Model: It performs slightly better than the GHR model in spacing prediction accuracy.

\item LSTM without PIDL without meta without pre-train Model: It performs slightly better than GHR and IDM models in spacing prediction but struggles with collision rate.

\item LSTM without PIDL without meta Model: It shows improvements in both spacing prediction and collision rate compared to the previous model.

\item LSTM without PIDL with meta Model: It further improves spacing prediction and collision rate by incorporating meta-learning methods.

\item LSTM with PIDL without meta Model: It achieves significant enhancements in reducing collision rate by utilizing PIDL methods.

\item LSTM with PIDL with meta (\textbf{MetaFollower}) Model: It outperforms all other models, showing the best performance in terms of spacing prediction and collision rate.
\end{itemize}

\begin{table*}[htbp]
\centering
\caption{Model  Performance: MSE of spacing and Collison rate }
\label{tab:results}
\begin{tabular}{@{}l|l|l@{}}
\toprule
Model & MSE of spacing                       & Collison rate \textperthousand ~  \\ \midrule
GHR   &  33.93                            & 0    \\
IDM   &  29.26                              & 0    \\
LSTM without PIDL without meta without pre-train & 26.67               & 55.82 (34)     \\
LSTM without PIDL without meta & 16.86               & 24.63 (15)     \\
LSTM without PIDL with meta & 14.66                & 18.06 (11)    \\
LSTM with PIDL without meta & 14.06               &   6.57 (4)  \\
LSTM with PIDL with meta (\textbf{MetaFollower}) &  \textbf{10.95 }               & \textbf{0}   \\

\bottomrule
\end{tabular}
\end{table*}

\section{Summary and Conclusion}
In this study, we propose a novel MetaFollower model for car-following behavior prediction, which, to the best of the authors' knowledge, is the first car-following model that leverages the knowledge of meta-learning considering both driver and temporal heterogeneity. The evaluation was conducted on the naturalistic driving dataset using two key metrics: MSE of spacing and collision rate. We compared its performance with six baseline models. Results demonstrated that the MetaFollower model outperformed baseline models in terms of both accuracy and safety. It achieved a lower MSE of spacing and a significantly reduced collision rate, indicating improved precision in predicting spacing and an enhanced ability to avoid collisions. The superior performance of the MetaFollower model can be attributed to its unique architecture and the integration of meta-learning techniques. By leveraging insights gathered from multiple drivers under the MAML framework, the MetaFollower model synthesizes a comprehensive understanding of car-following behavior and captures nuances of various driving styles. In conclusion, our study introduces the MetaFollower model as an effective solution for car-following behavior prediction which also highlights its potential for enhancing driver assistance systems and autonomous vehicle technologies. 

\section{Future Work}

Our research on the MetaFollower model for car-following behavior prediction opens up several avenues for future exploration and enhancement. Here are some potential directions for further investigation:
\begin{itemize}
    
    \item \textit{Data Augmentation}: Augmenting the training dataset with a wider range of driving scenarios and conditions could enhance the model's ability to generalize to diverse real-world situations. This could involve incorporating data from different geographic locations, weather conditions, and traffic patterns.
    
    \item \textit{Domain Extension}: While our study focuses on car-following behavior prediction, exploring the adaptability of the MetaFollower model to other driving tasks could be valuable. This could involve extending its capabilities to tasks such as merging, and trajectory forecasting.
    
    \item \textit{Real-time Implementation}: Investigating the feasibility of implementing the MetaFollower model in real-time systems, such as advanced driver assistance systems (ADAS) or autonomous vehicles, is crucial. This would involve optimizing the model's architecture and computational efficiency to meet the real-time constraints of such applications.
    
    \item  \textit{Collaborative Learning}: Investigating the potential for collaborative learning among multiple autonomous vehicles equipped with MetaFollower models could lead to enhanced cooperative behavior and further improve safety and efficiency in traffic.
\end{itemize}

\section{Acknowledgements}

This study was sponsored by Research on data-driven microscopic traffic simulation model for autonomous driving algorithm evaluation, Guangzhou. Basic and Applied Basic Research Project (SL2022A03J00083) and Guangzhou Municipal Science and Technology Project (2023A03J0011).

\section{Author contributions statement}
The authors confirm contribution to the paper as follows: study conception and design: Xianda Chen, Kehua Chen; data collection: Xuesong Wang, Meixin Zhu; analysis and interpretation of results: Xianda Chen, Kehua Chen; draft manuscript preparation: Xianda Chen, Kehua Chen. All authors reviewed the results and approved the final version of the manuscript.

\newpage
\bibliographystyle{trb}
\bibliography{trb_template}
\end{document}